\documentclass[11pt,a4paper]{article}
\usepackage[hyperref]{emnlp2020}
\usepackage{times}
\usepackage{latexsym}

\usepackage{microtype}

\aclfinalcopy % Uncomment this line for the final submission
\def\aclpaperid{2576} %  Enter the acl Paper ID here

\usepackage{times}

\usepackage{graphicx}
\usepackage{arydshln}
\usepackage{booktabs}

\usepackage{url}
\newcommand{\ignore}[1]{}

\usepackage{soul}
\usepackage{lipsum}
\newcommand{\X}{\mathcal{X}}

\usepackage{amsmath}
\usepackage{makecell}
\usepackage{soul}

\title{Resource-Enhanced Neural Model for Event Argument Extraction}

\author{Jie Ma\thanks{~ Indicates Equal Contribution.} ~~~~ Shuai Wang\footnotemark[1] ~~ \\ \textbf{Rishita Anubhai} ~~ \textbf{Miguel Ballesteros} ~~ \textbf{Yaser Al-Onaizan} \\ Amazon AI \\ {\tt \footnotesize{\{jieman, wshui, ranubhai, ballemig, onaizan\}@amazon.com}}}

\date{}

\begin{document}
\maketitle
\begin{abstract}
Event argument extraction (EAE) aims to identify the arguments of an event and classify the roles that those arguments play. Despite great efforts made in prior work, there remain many challenges: (1) Data scarcity. (2) Capturing the long-range dependency, specifically, the connection between an event trigger and a distant event argument. (3) Integrating event trigger information into candidate argument representation. For (1), we explore using unlabeled data in different ways. For (2), we propose to use a syntax-attending Transformer that can utilize dependency parses to guide the attention mechanism. For (3), we propose a trigger-aware sequence encoder with several types of trigger-dependent sequence representations. We also support argument extraction either from text annotated with gold entities or from plain text. Experiments on the English ACE2005 benchmark show that our approach achieves a new state-of-the-art.
\end{abstract}

\section{Introduction}
\label{sec:introduction}
\begin{figure}[t]
\begin{center}
\includegraphics[scale=0.28]{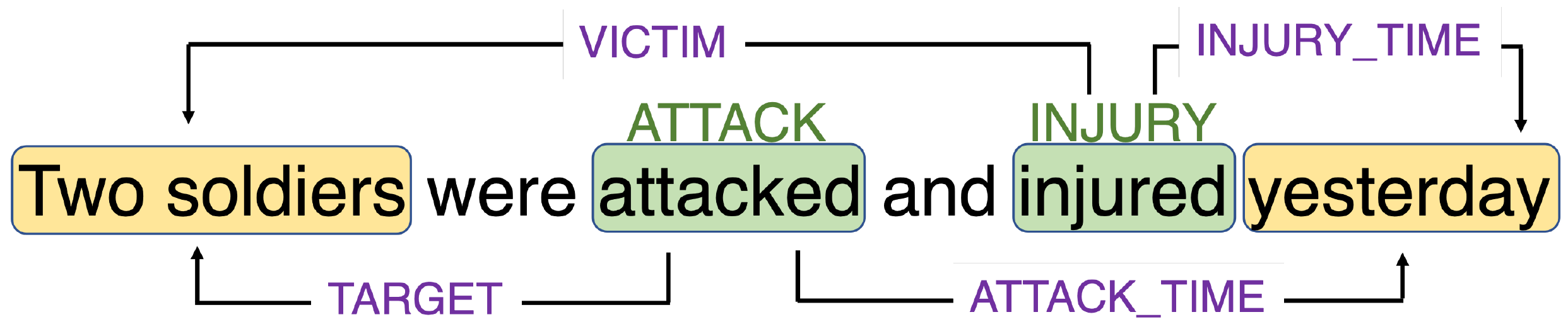}
\end{center}

\caption{Event examples: Green indicates triggers with their types. Yellow indicates arguments. An arrow indicates the role played by an argument in a trigger.}
\label{fig:example_event}
\end{figure}

Event argument extraction (EAE) aims to identify the entities that serve as arguments of an event and to classify the specific roles they play. As in Fig.~\ref{fig:example_event}, ``two soldiers'' and ``yesterday'' are \textit{arguments}, where the event triggers are ``attacked'' (with event type being ATTACK\footnote{Following ACE \url{https://www.ldc.upenn.edu/collaborations/past-projects/ace}})  and ``injured'' (event type INJURY). For the trigger ``attacked'', ``two soldiers'' plays the argument role \textit{Target} while ``yesterday'' plays the argument role \textit{Attack\_Time}. For the event trigger ``injured'', ``two soldiers'' and ``yesterday'' play the role \textit{Victim} and \textit{INJURY\_Time}, respectively. There has been significant work on event extraction (EE) ~\cite{liao2010using, hong2011using, li2013joint}, but the EAE task remains a challenge and has become the bottleneck for improving the overall performance of EE~\cite{wang2019hmeae}.\footnote{EAE has similarities with semantic role labeling. Event triggers are comparable to predicates in SRL and the roles in most SRL datasets have a standard convention of interpreting \textit{who} did \textit{what} to \textit{whom}. EAE has a custom taxonomy of roles by domain. We also use inspiration from the SRL body of work \cite{strubell2018linguistically, wang-etal-2019-best, he-etal-2017-deep, marcheggiani-titov-2017-encoding}.}

Supervised data for EAE is expensive and hence scarce. One possible solution is to use other available resources like unlabeled data. For that, (1) We use  BERT~\cite{devlin2018bert} as our model encoder which leverages a much larger unannotated corpus where semantic information is captured. Unlike%previous studies~\cite{wang2019hmeae, }
~\newcite{yang2019exploring} who added a final/prediction layer to BERT for argument extraction, we use BERT as token embedder and build a sequence of EAE task-specific components (Sec.~\ref{sec:solution}). (2) We use (unlabeled) in-domain data to adapt the BERT model parameters in a subsequent pretraining step as in \cite{gururangan2020don}.
This makes the encoder domain-aware. (3) We perform self-training to construct auto-labeled data (\textit{silver data}).

A crucial aspect for EAE is to integrate event trigger information into the learned representations. This is important because arguments are dependent on triggers, i.e., the same argument span plays completely different roles toward different triggers. An example is shown in Fig.~\ref{fig:example_event}, where ``two soldiers'' plays the role \textit{Target} for the event ATTACK and the role \textit{Victim} for INJURY. Different from existing work that relies on regular sequence encoders, we design a novel trigger-aware encoder which simultaneously learns four different types of trigger-informed sequence representations. %for candidate arguments. 

Capturing the long-range dependency is another important factor, e.g., the connection between an event trigger and a distant argument. Syntactic information could be useful in this case, as it could help bridge the gap from a word to another distant but highly related word~\cite{sha2018jointly,liu2018jointly, strubell2018linguistically}. We modify a Transformer \cite{devlin2018bert} by explicitly incorporating syntax via an attention layer driven by the dependency parse of the sequence. %(see Sec.\ref{subsec:syntax_encoder}) .

%Since arguments of an event are entities, entity mentions are very effective hints. 
We design our role-specific argument decoder to seamlessly accommodate both settings (with and without the availability of entities). We also tackle the \textit{role overlap problem} \cite{yang2019exploring} using a set of classifiers or taggers in our decoder. 

Our model achieves the new state-of-the-art on ACE2005 Events data~\cite{grishman2005nyu}.% for EAE.

%(see LISA for some related works)
% Motivation 1: data scarcity. Proposed and used solutions: (1) pretrained model BERT (2) External embedding (Double Emb) (3) Self-training (hard) (4) BERT MLM (5) MLM encoder and decoder joint pre-training. (6) Teacher-Student (soft)

%\section{Task Definition}
\section{Event Argument Extraction}
\label{sec:solution}
%This section shows the task setup for EAE. We then describe the key components of our model.   

%%%
%%\hl{(After Revision-Shuai)}
%%

\subsection{Task Setup}
\label{subsec:task_setup}
Consider a sequence $\X = \{x_1, ... x_i, ... x_T\}$ of $T$ tokens $x_t$. A span $x_{ij}$ =$\{x_i..x_j\}$ is a subsequence in $\X$. %starting with $x_i$ and ending with $x_{j-1}$
An event trigger $g$ is a span $x_{ab}$ indicating an event of type $y_g$, where $y_g$ belongs to a fixed set of pre-defined trigger types. Given a sequence-trigger pair ($\X$, $g$) as input, EAE has two goals: (1) Identify all \textit{argument} spans from $\X$ and (2) Classify the \textit{role} $r$ for each argument. In some settings, a set of entities is given (each entity is a span in $X$) and such entities are used as a candidate pool for arguments. For example, ``two soldiers'' and ``yesterday'' are candidate entities in Fig.~\ref{fig:example_event}.
% \shuaicomment{Key changes are highlighted in $Yellow$. Below is the original version without my changes. %Revision reason: $x_i$ and $x_j$ have not defined. Also revise some other statements and cut a bit.
% }

%%
% After Revision on May 31
%%

\subsection{Modeling Argument Extraction}
\label{subsec:model}
\begin{figure*}[t]
    \centering
    \includegraphics[scale=0.45]{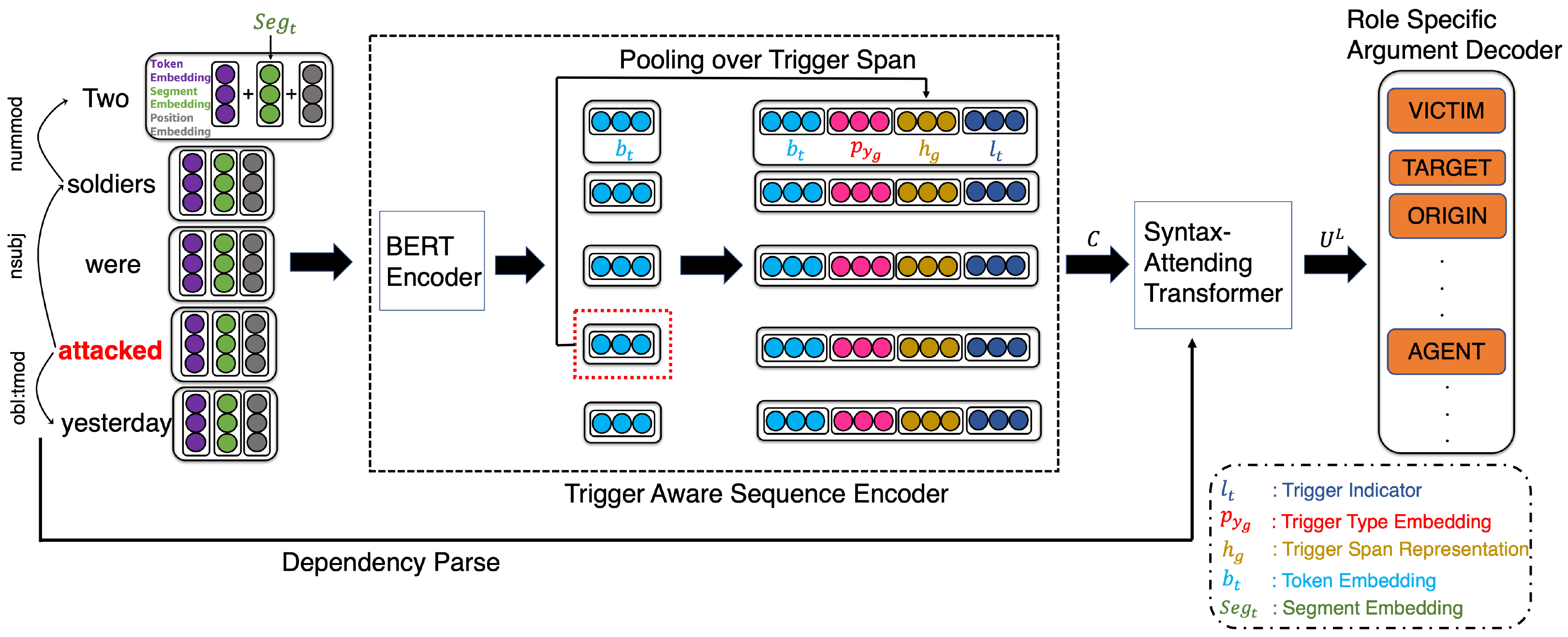}
    \caption{Model Architecture}
    \label{fig:model_figure}
\end{figure*}

Fig.~\ref{fig:model_figure} presents our model architecture with the following components: (1) trigger-aware sequence encoder, (2) syntax-attending Transformer and (3) role-specific argument decoder.

%  \hl{(After Revision-Shuai)}

\paragraph{Trigger-Aware Sequence Encoder:} 
\label{subsec:trigger_aware_encoder}
%~\newcite{wang2019hmeae} and \newcite{liu2018jointly} relied on encoding the input sequence $\X$ using a vanilla BERT-based sequence encoder without using any trigger information. The resulting independent representations were then used alongside the trigger representations. 

% \begin{figure}[!ht]
% \begin{center}
% \includegraphics[scale=0.7]{trigger_aware_example_no_entities.png}
% \end{center}

% \caption{Example where the same span encodes different argument information for two triggers. For brevity, we do not indicate unrelated roles and arguments in this example. Rest as in Fig. \ref{fig:example_event}.}
% \label{fig:trig_aware_event}
% \end{figure}

This encoder is designed to distinguish candidate arguments conditioned on different triggers. Note a span may encode different argument information for two triggers, for example, in Fig. \ref{fig:example_event}, ``Two soldiers" plays the role of \textit{Target} for the ATTACK event and \textit{Victim} for the INJURY event. In order to model this, our encoder uses BERT to embed input tokens, where the BERT embedding of token $x_t$ is denoted as $b_t$. A segment (0/1) embedding $seg_{t}$ for each token $x_t$ indicating whether $x_t$ belongs to the trigger or not \cite[inter-alia]{logeswaran-etal-2019-zero} is used, which is added up with token embedding and position embedding as input to BERT (Fig.\ref{fig:model_figure}).
%Each token $x_t$ is added to a (0/1) embedding $seg_{t}$ indicating whether $x_t$ belongs to the trigger or not \cite[inter-alia]{logeswaran-etal-2019-zero}. %; see Fig.\ref{fig:model_figure}, the module to the left of BERT.
%and our model learns this as its embeddings, i.e., indicator embeddings. Specifically, the segment embeddings ($seg_{t}$) as an input to BERT are used as our indicator embeddings (see Fig~\ref{fig:model_figure}).
The encoder then concatenates the following learned representations\footnote{(2) and (3) are randomly initialized. They will be jointly learned during model training.} for each token: (1) A trigger representation $h_g$ by max pooling over BERT embeddings of the tokens in trigger $g$; (2) A trigger type embedding $p_{y_g}$ for $y_g$; (3) A trigger indicator (0/1) embedding $l_t$, indicating whether $x_t$ belongs to the trigger or not.\footnote{While both $l_t$ and $seg_t$ are used for indicating whether a token $x_t$ belongs to a trigger, the difference is that $l_t$ is used to encode such information after obtaining BERT outputs and $seg_t$ is used as an input to BERT. They are of different sizes.}%; they are of different sizes.} %\footnote{The segment embedding $seg_t$ is of different size since it is added to the BERT input vector both $l_t$ and $seg_t$ represent the same information.}
%learned from the binary indicator $m_t$ defined above
(4) A token embedding $b_t$. This results in a trigger-aware representation $c_t$ for each token where $c_t = Concat(b_t;p_{y_g};l_t;h_g)$ and $C$ for the whole sequence with T tokens.

%%%
%   Temporarily comment out the original section 2.2 and move it to the end
%%%

%%
% After Revision on May 31
%%
%\hl{(After Revision-Shuai}
\paragraph{Syntax-Attending Transformer:}
\label{syntax-att}
Dependency parsing has been used as a feature to improve EE ~\cite{sha2018jointly,liu2018jointly}. Inspired by ~\newcite{strubell2018linguistically}, we utilize dependency parses\footnote{We use Stanford Parser \url{https://nlp.stanford.edu/software/lex-parser.shtml}} by modifying an attention head for each layer in a Transformer. Note that this Transformer is different from the BERT component, as this Transformer aims to capture long-range dependency on top of the trigger-aware representations learned from our sequence encoder. The output $C$ from our encoder now will be the input of this Transformer, which will go through $L$ layers of the modified syntax-attending Transformers. Each of these is assumed to have $N$ self-attention heads. For each layer $l$, we modify one of these $N$ heads to be a dependency-based attention head (call d-head) with output $H^{l}$: 

\begin{equation}
\label{eq:depheadoutput}
	 H^{l} = W_2([W_1U^{(l-1)};A^{l}]),
\end{equation}
where
\begin{equation}
\label{eq:depheadattention}
	A^{l} = Attention(W_Q^{l}Q^{l}, W_K^{l}K^{l}, W_V^{l}V^{l}).
\end{equation}
$Q^{l}$, $K^{l}$ and $V^{l}$ are query, key, and value representations, and $W_{*}$ are learning parameters. $U^{l} = \{u_1^l .. u_T^l \}$ is the layer-$l$ output of our Transformer and $U^{0} = C$.
Eq. \ref{eq:depheadattention} uses the scaled-dot product attention \cite{vaswani2017attention}. The difference of the d-head compared to other heads is that its keys $K$ and values $V$ are constructed differently. For each token $x_i$, valid keys and values are restricted to all tokens $x_j$ such that $x_i$ and $x_j$ have an edge between them\footnote{$x_i$ may represent a subword unit. We assume all subwords of a word in the dependency parse inherit the head and children from the parent word.
} in the dependency parse of the sequence $\X$. This makes every $a_t^{l} \in A^{l}$ a weighted attention sum over the neighbor\footnote{Head and children.} values $v_j^{l}$ of the token $x_i$ in the dependency parse. We then concatenate this $a_t^{l}$ and the token's own representation $u_t^l$ projected linearly. Finally, this is projected back to the same dimensions as the outputs of the other $N-1$ attention heads. By concatenating their outputs, our model captures both syntax-informed and global-attending information. The final output from our Transformer component is $U^{L} = {u^{L}_1, u^{L}_2 ..u^{L}_T}$.

%%
% After Revision on May 31
%%
%\hl{(After Revision-Shuai}[WIP]
\paragraph{Role-Specific Argument Decoder:}
\label{subsec:decoder}

We consider two settings: (1) with and (2) without entities. When entities are provided, they are used to form candidates for arguments; when they are not provided, our model infers arguments from plain text.

For (1), we assume that all arguments are entities but the vice versa is not true. So, we treat all entity spans, within a fixed sentence window around the trigger $g$, as candidate arguments. An entity representation is formed by pooling $u^L_t$ for all tokens $x_t$ in the entity span. Note that, since the encoder is trigger-aware, this representation is already conditioned on ($\X$, $g$) for role classification. Commonly used datasets like ACE2005 have a 10\% \textit{role overlap problem} ~\cite{yang2019exploring}. Concretely, consider a sentence like ``The suicide bomber died in the blast he set off". Here, ``suicide bomber" plays two distinct roles \textit{Attacker} and \textit{Victim} for the same trigger    ``blast" that denotes an ATTACK event. Hence, we perform role classification for every role independently (as a multi-label classification problem), using a set of classifiers, where each classifier handles one particular role, i.e., role-specific (such as the \textit{VICTIM}, \textit{TARGET} or \textit{ORIGIN} as orange shown in Fig.~\ref{fig:model_figure}). We thus call this decoder role-specific argument decoder. 

More specifically, we use one binary classifier per role permissible for current trigger type on this entity representation. The outcome of the classifier for role $r$ determines whether this entity plays the role $r$ for the current trigger or not. 

For (2), in the absence of entities we have no candidate spans for arguments. Using final layer output of syntax-attending Transformer, we predict a sequence of BIO tags with one sequence tagger per role.\footnote{Each token is tagged with B, I or O, indicating a token is at the beginning, inside of, or outside an argument span. Here we replace the conventional multi-class BIO tagger with a set of role-wise taggers. So tokens play role $r$ have the tags B and I from the role-specific tagger for $r$.  %(Fig. ~\ref{fig:model_figure}). %to handle the role overlap problem mentioned above.
} So in this setting the role-specific argument decoder comprises a set of sequence taggers.

%Using the syntax-attending Transformer output, we predict a sequence of BIO tags with one sequence tagger per role.
%Using the final syntax-attending layer, we predict a sequence of BIO tags with one sequence tagger per role.

%Using the syntax-attending Transformer output, a set of role-based taggers make BIO predictions.

% \footnote{Each token is tagged with B, I or O. Words outside argument spans have the tag O, and words at the beginning and inside of argument spans playing role $r$ have the tags B and I from the role decoder for $r$. Here also, we replace the conventional multi-class BIO tagger with a set of role-wise taggers. %(Fig. ~\ref{fig:model_figure}). %to handle the role overlap problem mentioned above.
% }

%% delete this (old version) - shuai commented
% \paragraph{Support Gold Entity and End-to-End Setting:} As discussed above, our proposed model can be either based on gold (ground-truth) entity mentions or not, which is not the case for most existing models. For them, they mostly assumed the entities are known. For us, we are not limited to but also support this setting, where our role-specific taggers can reduce to multi-label classifiers as a flexible change, simply resulting in a model variant. More importantly, while no gold entities are given (which is a more realistic case in real world data), our model will also work as in this end-to-end EAE setting. Here we want to highlight that such modeling flexibility and extensibility are also advantages of our proposed solution.

\subsection{Training Regimes for Data Scarcity}
\label{subsec:other_modeling_parts}
%We experiment with the following training regimes. 

\paragraph{Domain-adaptive pretraining:}
An additional phase of in-domain pretraining has been shown to be effective for downstream tasks ~\cite{gururangan2020don}. Based on this, we perform a second phase of domain-adaptive pretraining with both BERT losses before fine-tuning the BERT encoder. %Specifically, we use in-domain unlabeled data (Sec. ~\ref{subsec:data}) and further train a BERT\textsubscript{BASE} checkpoint which was trained on English book corpus and Wikipedia.

\paragraph{Self-training:} 
For self-training~\cite[inter-alia]{chapelle2009semi,scudder1965probability}, we first train our model based on gold data. Next, we use that model to tag unlabeled data and get a much larger but noisy silver dataset (Sec.~\ref{subsec:data}). We then train a new version of our model on the silver dataset; the resulting model is later fine-tuned on the gold data.

\paragraph{Auxiliary tasks:} Although trigger detection is not the focus of this work, we model it as an auxiliary task to help EAE. We share the BERT encoder (Sec.~\ref{subsec:model}) for both tasks. The trigger detection task uses the standard sequence tagging decoder for BERT \cite{devlin2018bert}. The intuition here is to improve (1) the representation of the shared BERT component and (2) trigger representation, by performing trigger detection.
% The intuition here is to improve the understanding of the shared encoder by predicting triggers. 
% Note, since our main task is EAE,  our model is not tuned for best trigger predictions scores.
% Our model is not tuned for best trigger prediction scores, but this setup could be further used to jointly perform both tasks.

\section{Experiments}
\label{sec:experiments}
%This section shows our results, comparison with prior work and ablation analysis experiments.
% %\miguelcomment{add a paragraph explaining what the reader is going to read next}
% This section illustrates the evaluation of our proposed approach. 
% Data details, %evaluation criteria, 
% candidate models for comparisons\miguelcomment{comparison with prior work}, evaluation results, and the ablation analyses, would \miguelcomment{would?} be reported.
% \miguelcomment{how about this?: This Section shows our results, comparison with prior work and ablation analysis experiments.}
\paragraph{Data and Tools:}
\label{subsec:data}
%Table X shows the results of all candidate models on ACE Event 2005 dataset.
We use the ACE-2005 English Event data~\cite{grishman2005nyu}.\footnote{Standard splits \cite{li2013joint}: 529 documents (14,385 sentences) are used for training, 30 documents (813 sentences) for development, and 40 documents (632 sentences) for test.} %\shuaicomment{Todo: add a link}%\miguelcomment{where is the split found, can we put the website/paper here?}
For self-training and domain-adaptive pretraining, we randomly sample 50k documents (626k sentences) from Gigaword\footnote{https://catalog.ldc.upenn.edu/LDC2011T07} to construct silver data. We use Stanford CoreNLP software\footnote{https://stanfordnlp.github.io/CoreNLP/} for tokenization, sentence segmentation and dependency parsing.
%\miguelcomment{do PLMEE authors use the same split? do we mention this somewhere?}

\paragraph{Training Setup:}
We use 50 dimensions for trigger indicator and trigger type embedding. We use 2 ($L=2$) layers for the syntax-attending Transformer with 2 ($N=2$) attention heads, dropout of 0.1. When entities are available, we only consider entities in the same sentence as the trigger as candidates for argument extraction. During training, We use Adam~\cite{kingma2014adam} as optimizer and batch size of 32 for both main task EAE and the auxiliary task of trigger detection; we alternate between batches of main task and auxiliary task with probabilities of 0.9 and 0.1, respectively. We early stop training if performance on the development set does not improve after 20 epochs. All model parameters are fine-tuned during training. For BERT pretraining, we use the same setting as in ~\cite{devlin2018bert} but with an initial learning rate of 1e-5. We stop pretraining after 10k steps. In order to obtain reliably predicted triggers as input for EAE, we trained a five-model ensemble trigger detection system following \newcite{wadden2019entity}.\footnote{Since trigger detection is not our main task and improving it is not the focus of this work, its results are not for comparison and thus excluded from the main result tables. As a reference, our five-model ensemble achieves 73.88 F1 score in the trigger classification task on ACE2005 Event test set.} % which is optimized to achieve best performance on trigger detection task.

\paragraph{Results and Analyses:}
% Table \ref{restable} reports the experimental results of all model for argument identification (AI) or role classification (CL), from which we made the following observations: (analyses to be added). 

Table~\ref{restable} shows the results in two experimental settings: with and without entities. In the setting with entities, we use gold entities as in prior work.
% Evaluation metrics are precision (P), recall (R) and F1 score (F) on argument identification (AI) and role classification (RC).
%where AI is an intermediate step without evaluating the role types and RC is the final measure considering both the correctness of argument assignment, argument boundary detection and type classification. 
We have the following observations: (1) Our model achieves the best results ever reported in both experimental settings on RC (overall F1 scores). (2) Our model does not achieve the highest scores on AI. It seems however that our model is able to bridge the gap given that in order to achieve good results in RC you need AI, so it couples these two mutually affected sub-tasks closer to each other.
(3) Self-training leads to gains of 1 F1 point both with and without entities. (4) Domain-adaptive pretraining shows small improvements on both AI and RC (less that self-training). There are two possible reasons. First, Gigaword is news while ACE is not only news; we only adapted part of the domains. Second, even though we used a small learning rate during pretraining, 50k unlabeled documents is a small amount for pretraining. %It is possible that after our pretraining the advantage of original BERT also diminished. 

%%
%	Citations as model names (and with old baselines removed)
%%

\begin{table}[t]
%\begin{table}[!ht]
\centering
\scalebox{0.64}{
\begin{tabular}{c|ccc|ccc}
\toprule
% \textbf{Experiment} &  \textbf{Argument (F1)} & \textbf{Trigger (F1)} \\
\multicolumn{1}{c}{\textbf{Model}} &  \multicolumn{3}{c}{\textbf{\makecell{Argument \\ Identification (AI)}}} &  \multicolumn{3}{c}{\textbf{\makecell{Role \\ Classification (RC)}}} \\
%\midrule
 \multicolumn{1}{c}{} &  \multicolumn{1}{c}{P} & R & \multicolumn{1}{c}{F1}  &  \multicolumn{1}{c}{P} & R & F1 \\
%  \multicolumn{7}{c}{\textbf{Gold Entities Given}} \\
\midrule  
%\small{\cite{liao2010using}}  & {50.9} & 49.7 & 50.3 & 45.1 & 44.1 & 44.6 \\
%Cross-Entity & 53.4 & 52.9 & 53.1 & 51.6 & 45.5 & 48.3 \\
%JointBeam & 69.8 & 47.9 & 56.8 & 64.7 &44.4 & 52.7\\
\cite{yubo2015event} & 68.8 & 51.9 & 59.1 & 62.2 & 46.9 & 53.5 \\
\cite{nguyen2016joint} & 61.4 & 64.2 & 62.8 & 54.2 & 56.7 & 55.4 \\
% JREE  &  55.4 & 69.3 \\
\cite{sha2016rbpb}  & 63.2 & 59.4 & 61.2 & 54.1 & 53.5 & 53.8 \\
\cite{sha2018jointly} & 71.3 & 64.5 & 67.7 & 66.2 & 52.8 & 58.7 \\
\cite{yang2019exploring} &  71.4 & 60.1 & 65.3 & 62.3 & 54.2 & 58.0 \\
\cite{wang2019hmeae} & - & - & - & 62.2 & 56.6 & 59.3 \\
\cite{liu2018jointly}  & \textbf{71.4} & \textbf{65.6} & \textbf{68.4} & \textbf{66.8} & 54.9 & 60.3 \\
\midrule
Ours  &  64.8 & 63.7 & 64.2 & 61.1 & 60.6 & 60.8\\
Ours + Pretraining & 65.8 & 62.9 & 64.3 & 62.3 & 60.0 & 61.1  \\
Ours  + Self Training & 64.5 & 65.0 & 64.7 & 61.1 & \textbf{62.3} & \textbf{61.7}  \\
%\midrule
%(w/o Gold Entity) &  P R F   &  P R F \\
\midrule
\midrule
% Joint IE$^{\dagger}$     &  .. & .. & .. & 60.5 & 39.6 & 47.9 \\
%JointEntityEvent$^{\dagger}$  &  73.7 & 38.5 & 50.6 & \textbf{70.6} & 36.9 & 48.4 \\
\cite{sha2018jointly}$^{\dagger}$    &  - & - & 57.2 &   -  & - &50.1 \\
%GAIL-W2V$^{\dagger}$    &  62.3 & 48.2 & 54.3 & 61.7 & 44.8 & 51.9 \\
\cite{zhang2019joint}$^{\dagger}$    &  63.3 & 48.7 & 55.1 & \textbf{61.6} & 45.7 & 52.4 \\
\cite{nguyen2019one}$^{\dagger}$   &  \textbf{59.9} & \textbf{59.8} & \textbf{59.9} & 52.1 & 52.1 & 52.1 \\
\cite{wadden2019entity}$^{\dagger}$   &  - & - & 55.4 & - & - & 52.5 \\
\cite{zhang2020question}$^{\dagger}$ &   - & - & -  &  54.5 & 52.4 & 53.4 \\
\midrule
Ours  $^{\dagger}$  &  55.6 & 57.9 & 56.7 & 53.0 & 55.7 & 54.3\\
Ours  + Pretraining $^{\dagger}$ & 56.3 & 58.1 & 57.2 & 53.5 & \textbf{55.8} & 54.6  \\
Ours  + Self Training $^{\dagger}$ & 58.4 & 56.9 & 57.6 & 56.0 & 54.8 & \textbf{55.3}  \\
% Other more variants  &  (pending) & (pending)\\
% Trigger task basic &  - & around 79  \\
% Argument task basic &   60.38 \scriptsize{$\pm$ 1.05} & - \\
% Argument task + ind + max + mean &   61.62 \scriptsize{$\pm$ 0.94} & - \\
% \midrule
% Multi-task (current best, ind + mean + max + 2 decoders) &   62.48 \scriptsize{$\pm$ 0.38} & 79.54 \\
\bottomrule
\end{tabular}
}
\caption{Experimental results. %The results of TASDE are based on 5 models (5 different runs). %\miguelcomment{can we run 5 models and show mean and standard deviation?}\shuaicomment{Yes. We are going to add it} 
$\dagger$ indicates a model does not use gold entities. 
%The models for comparisons: {Cross-Event}~\cite{liao2010using}, {Cross-Entity}~\cite{hong2011using}, {JointBeam}~\cite{li2013joint}, {DMCNN}~\cite{yubo2015event}, {JRNN}~\cite{nguyen2016joint}, {RBPB}~\cite{sha2016rbpb}, {dbRNN}~\cite{sha2018jointly}, {PLMEE}~\cite{yang2019exploring}, {HMEAE}~\cite{wang2019hmeae}, {JMEE}~\cite{liu2018jointly}, 
%%{JointIE}~\cite{li2013joint}, 
%JointEntityEvent~\cite{yang2016joint}, GAIL~\cite{zhang2019joint}, JOINT3EE~\cite{nguyen2019one}, DYGIE++~\cite{wadden2019entity} and QA-SL~\cite{zhang2020question}. 
Ours show the mean of 5 random seeds. P refers to precision and R refers to recall.
%HMEAE, PLMEE, QA-SL are based on BERT.
%\miguelcomment{maybe add the citations in the table directly like we did here: https://arxiv.org/pdf/2004.04295.pdf, also the numbers are too close to each other because there are no column separators, it is difficult to distinguish}\shuaicomment{todo: address this comment.}
%\miguelcomment{If some models are papers, please cite them. If some models are described in the paper, please make sure to refer to the corresponding section here}
}
\label{restable}
\end{table}

\paragraph{Ablation Study:}
We ablate each of the components and show results in Table~\ref{ablation}. We observe that (1) All components help. We can see the performance gain of each component in the settings of with and without entities.  (2) The overall trigger-aware sequence encoder leads to $\sim$1.5 F1 points gain in both settings. (3) The use of the auxiliary task and the syntax-encoder improve by another $\sim$1 F1 points. %\shuaicomment{We may explain a bit more here, with more space available now.}
%The ablation study is reported in Table 2.
%\miguelcomment{and what are you ablating?}  

\begin{table}[!ht]
\centering
% \scalebox{0.7}{ % submssion
\scalebox{0.78}{ % submssion
\begin{tabular}{l|c|cc}
\toprule
% \textbf{Experiment} &  \textbf{Argument (F1)} & \textbf{Trigger (F1)} \\
% &  \textbf{Entity } & (TBF) \\
% \textbf{Model} &  \textbf{With entities} & \textbf{Without entities} \\ % submssion
\textbf{Model} &  \textbf{w/ entities} & \textbf{w/o entities} \\
% \textbf{} &  \textbf{Identification} & \textbf{Classification} \\
\midrule
Argument (Single Task) & 68.1 & 62.1 \\
Argument + TI & 68.5 & 62.5 \\
Argument + TI + TT  & 69.1 & 63.3 \\
Argument + TI + TT  + TE & 69.6 &  63.5 \\
Arg. + Tri. (Auxiliary) & 70.2 & 64.2 \\
Arg. + Tri. (Auxiliary) + Syntax. & 70.8 & 64.6 \\
% \midrule
% TASDE (our model)  &  (pending) & (pending)\\
% Other more variants  &  (pending) & (pending)\\
% Trigger task basic &  - & around 79  \\
% Argument task basic &   60.38 \scriptsize{$\pm$ 1.05} & - \\
% Argument task + ind + max + mean &   61.62 \scriptsize{$\pm$ 0.94} & - \\
% \midrule
% Multi-task (current best, ind + mean + max + 2 decoders) &   62.48 \scriptsize{$\pm$ 0.38} & 79.54 \\f
\bottomrule
\end{tabular}
}
\caption{Ablation analysis of our model on development set with gold trigger. TT = Trigger Type. TI = Trigger Indicator. TE = Trigger Embedding. Tri. = Trigger. Arg. = Argument + TI + TT + TE. Auxiliary indicates that trigger detection here is an auxiliary task. w/ and w/o entities mean with and with out entities provided. Results are the mean over 5 random seeds.
%\shuaicomment{we need to add standard deviations to both table 1 and table 2. Previously also suggested by Miguel. }
}
\label{ablation}
\end{table}
%($Arg. + TI + TT + TE$) 
 %They both strength the representation learning, where the former focuses on triggers and the latter focuses on long-range relations between triggers and arguments. %strength the representation learning of triggers and arguments.

\section{Related Work}
\label{sec:related_work}
Event Argument Exaction (EAE) is an important task in Event Extraction (EE). Early studies designed lexical, contextual or syntactical features to tackle the EE problem~\cite{ji2008refining,liao2010using,hong2011using, li2013joint}.
Later on, neural networks~\cite{yubo2015event,sha2016rbpb,nguyen2016joint,sha2018jointly,liu2018jointly,yang2019exploring, wang2019hmeae} demonstrated their effectiveness in representation learning without manual feature engineering. Our proposed model belongs to the latter category.
 
Here we present and discuss the most related studies to our work. ~\newcite{yang2019exploring} used a pre-trained model with a state-machine based span boundary detector. They used heuristics to resolve final span boundaries. ~\newcite{wang2019hmeae}
%used a pre-trained model in a state-machine based span boundary learning fashion, with the involvement of some heuristic designs. \rishitacomment{I would shorten and split it in summary with mention of what we think are the 'heuristic designs': "used a pre-trained model with a state-machine based span boundary detector. They used heuristics to resolve to final span boundaries."} 
%~\newcite{wang2019hmeae} %\rishitacomment{Why is this not in the list of neural models in first paragraph?}
also used a pre-trained model together with a hand-crafted conceptual hierarchy. %Their solutions both need human designs \rishitacomment{Do you mean manual feature-engineering? 'Human designs' is a bit vague.} while ours do not require. 
Our approach does not need the design of such heuristics or conceptual hierarchy. In terms of modeling, their approaches used regular BERT as their encoders%to generate the whole event sequence%and then add a prediction layer on top of it
, where the argument representations are not explicitly conditioned on triggers. %The representations from BERT encoder are then past to their \rishitacomment{which decoders, sequence taggers? Maybe call that?}decoders for argument extraction. %\rishitacomment{That is only their encoders. I assume you want to say sequence taggers on top of BERT for them?}. 
In contrast, our encoder is enhanced by providing more trigger-oriented information and BERT is only used as one part of it, which results in a trigger-aware sequence encoder. %Along with other sophisticated modeling units, our approach can better 
This allows us to better model interactions between arguments and triggers.
~\newcite{liu2018jointly} added a GCN layer to integrate the syntactic information into a neural model. Different from their solution, we encode the syntax jointly with attention mechanism, simplifying the learning, making it more efficient, and achieving better results. Finally, no prior work has deeply studied the data scarcity issue in EAE, while we exploit several techniques to tackle it in this work.
%\rishitacomment{One thing to add here is a comparison to dep parse papers and say how they use it at inference too. You use the information but only at training time, hence taking no inference cost of running a dep parser. This is quite strong I think in comparison.}

% (

%~\cite{wang2019hmeae} 

%~\cite{wang2019hmeae} does not consider addressing the role overlapping problem. ~\cite{yang2019exploring} used a pre-trained model in a relatively complex state-machine based span boundary learning fashion, which aims to extract start and end tokens with the involvement of heuristic designs. 
%
%A previous work JMEE also uses syntax, but in a complicated way.

\section{Conclusion}
We present a new model which provides the best results in the EAE task. The model can generate trigger-aware argument representations, incorporate syntactic information (via dependency parses), and handle the role overlapping problem with role-specific argument decoder. We also experiment with some methods to address the data scarcity issue. Experimental results show the effectiveness of our proposed approaches.

%Experimental results demonstrate the effectiveness of our approach.

%We also addressd the learning gap/discrepancy between pre-trained and newly-trained components.

\section*{Acknowledgments}
We would like to thank the anonymous reviewers for their comments and suggestions.

\bibliography{anthology,emnlp2020}
\bibliographystyle{acl_natbib}

\end{document}